\documentclass{article}

   \usepackage[final, nonatbib]{neurips_2024}

\usepackage[utf8]{inputenc} 
\usepackage[T1]{fontenc}    
\usepackage{hyperref}       
\usepackage{url}            
\usepackage{booktabs}       
\usepackage{amsfonts}       
\usepackage{nicefrac}       
\usepackage{microtype}      
\usepackage{xcolor}         
\usepackage{amsmath}
\usepackage{graphicx}
\usepackage{calc}
\usepackage{xcolor}
\usepackage{tcolorbox}
\usepackage{fancyhdr}

\usepackage{afterpage}

\usepackage{geometry}

\fancypagestyle{firstpagestyle}{
    \fancyhf{} 
    \fancyfoot[C]{\vspace{-0.1cm} 
    \rule{0.8\linewidth}{0.4pt}\\ 
    This is a preprint and current version under peer review.}
}

\title{Towards Automated Patent Workflows: AI-Orchestrated Multi-Agent Framework for Intellectual Property Management and Analysis}

\author{
  Sakhinana Sagar Srinivas$^{1}$, Vijay Sri Vaikunth$^{2}$, \textbf{Venkataramana Runkana$^{1}$}\\
  TCS Research$^{1}$, IIT–Palakkad$^{2}$ \\
  \texttt{sagar.sakhinana@tcs.com}, \texttt{112101060@smail.iitpkd.ac.in}, \texttt{venkat.runkana@tcs.com}
}

\begin{document}

\maketitle

\vspace{-4mm}
\begin{abstract}
\textit{“Patents are the currency of innovation, and like any currency, they need to be managed and protected”} (Gavin Potenza). Patents, as legal documents that secure intellectual property rights, play a critical role in technological innovation. The growing complexity of patent documents and the surge in patent applications have created a need for automated solutions in patent analysis. In this work, we present \textbf{PatExpert}, an autonomous multi-agent conversational framework designed to streamline and optimize patent-related tasks. The framework consists of a meta-agent that coordinates task-specific expert agents for various patent-related tasks and a critique agent for error handling and feedback provision. The meta-agent orchestrates specialized expert agents, each fine-tuned for specific tasks such as patent classification, acceptance, claim generation, abstractive summarization, multi-patent analysis, and scientific hypothesis generation. For multi-patent analysis, the framework incorporates advanced methods like Graph Retrieval-Augmented Generation (GRAG) to enhance response accuracy and relevance by combining semantic similarity with knowledge graphs. Error handling is managed by critique agents (Gold-LLM-as-a-Judge and Reward-LLM-as-a-Judge), which evaluate output responses for accuracy and provide iterative feedback. The framework also prioritizes explainability, ensuring transparent justifications for decisions made during patent analysis. Its comprehensive capabilities make it a valuable tool for automating complex patent workflows, enhancing efficiency, accuracy, and compliance in patent-related tasks. Empirical evidence demonstrates significant improvements in patent processing tasks, concluding that the framework offers a robust solution for automating and optimizing patent analysis.
\vspace{-4mm}
\end{abstract}

\section{Introduction} 
\begin{figure}[!ht]
\vspace{-2mm}
\centering
\resizebox{0.775\linewidth}{!}{ 
\hspace*{0mm}\includegraphics[keepaspectratio,trim=0.0cm 0cm 0cm 0.15cm,clip]{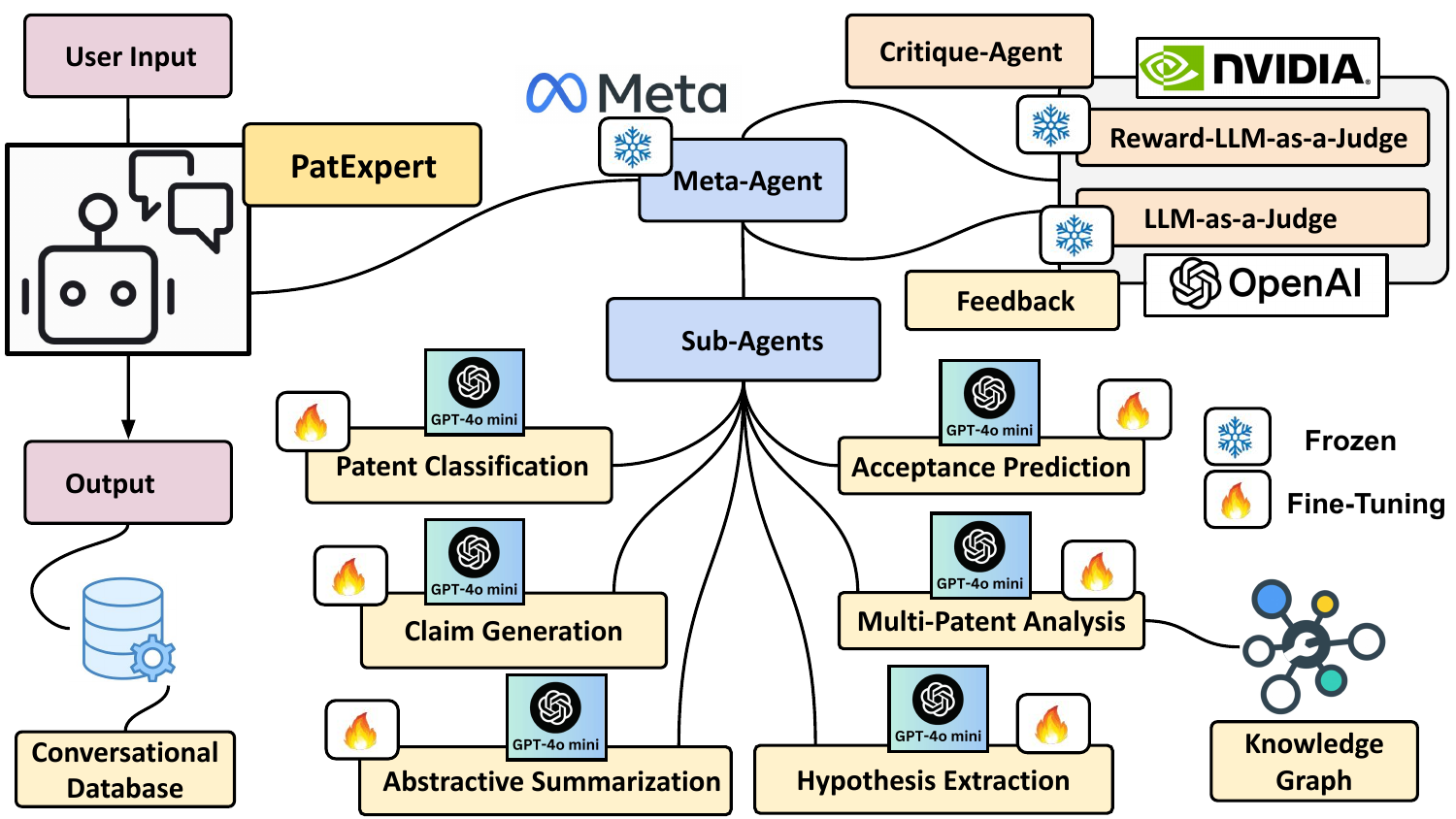} 
}
\vspace{-2mm}
\caption{
The figure shows the architecture of the multi-agent conversational framework, \textbf{PatExpert}. The meta-agent oversees various sub-agents, each responsible for specialized tasks such as patent classification, acceptance prediction, claim generation, abstractive summarization, hypothesis generation, and multi-patent analysis. User input flows through the meta-agent (which utilizes Meta-Llama-3.1-405B to interpret user queries), delegating tasks to the relevant sub-agents (using fine-tuned GPT-4o mini) to provide accurate responses. Critique agents, including the Reward-LLM-as-a-Judge (Nvidia Nemotron-4-340B-Reward) and LLM-as-a-Judge (OpenAI GPT-4o), evaluate the outputs for accuracy. These agents provide critical feedback to refine responses, ensuring that the framework adheres to high standards of precision, compliance, and quality in patent-related tasks. The knowledge graph stores structured information extracted from patents, while the conversational database holds historical interactions, helping the framework maintain context and continuity in multi-turn queries. The knowledge graph enables efficient semantic retrieval across multiple patents for ODQA tasks, improving response accuracy. In summary, \textbf{PatExpert} handles complex patent-related tasks and generates accurate, coherent responses through its structured, multi-agent framework.
}
\label{fig:figure1}
\vspace{-3mm}
\end{figure}

\vspace{-3mm}
\textit{“An invention is something that was `impossible' up to then; that's why governments grant patents.”} (Robert A. Heinlein). A patent is a legal document that grants the inventor exclusive rights to make, use, and sell their invention for a specified period in exchange for public disclosure of the invention's details. Patent documents are essential for securing intellectual property rights and serve as a public record of technological innovation. While these documents may vary slightly by jurisdiction (e.g., USPTO, JPO, EPO), they generally consist of several key sections that ensure both legal protection and clear communication of the invention. A patent typically includes a title, abstract, background, summary, detailed description, claims, and illustrations. The title highlights the invention's main innovation, while the abstract provides a brief overview of the invention, focusing on its purpose, key features, and potential applications. The background section outlines existing solutions and the technical challenges the invention addresses. The detailed description provides in-depth technical specifications, including various embodiments (specific versions or implementations of the invention), processes (methods or procedures involved in making or using the invention), and use cases (practical applications or scenarios where the invention could be applied), explaining how the invention addresses the identified challenges. The claims accurately determine the scope of legal protection, outlining the legal boundaries of coverage. These claims can be independent, broadly covering the core features, or dependent, refining the scope by specifying particular embodiments. Other important sections include patent classification codes, citations to prior art, and illustrations accompanied by detailed descriptions that visually depict key aspects of the invention, aiding in the understanding of complex technical details. Collectively, these main sections ensure that the patent document meets legal requirements, communicates the invention effectively, and provides enforceable rights to the inventor. This emphasizes the necessity for precision and clarity in patent drafting to ensure it withstands legal scrutiny and future challenges. In recent times, Large Language Models (LLMs), such as OpenAI GPT-4\cite{achiam2023gpt} and Google Gemini\cite{reid2024gemini}, have excelled in natural language processing tasks due to their proficiency in pattern recognition, contextual understanding, and generating coherent language based on learned data distributions. However, their application to patent-related tasks remains underexplored. Patents, as complex legal documents that protect intellectual property, combine technical information, requiring a deep understanding of the subject matter and precise language to describe complex, domain-specific concepts. Additionally, the rise in patent applications and the increasing difficulty of manual processing have created a growing need for LLMs to automate patent-related workflows. LLMs can enhance patent analysis, knowledge extraction, and document generation, transforming both the analysis and generation of patent content. In patent analysis, LLMs excel at automating the categorization of patents into relevant classes and subclasses based on their subject matter and technological field. For quality assessment, LLMs can evaluate patent novelty by comparing new patents against prior art, estimate the likelihood of approval based on historical data, and predict potential litigation by analyzing claim strength and market relevance. In multilingual translation, they maintain precise technical meanings across languages. In open-domain question answering (ODQA) tasks, LLMs using the Retrieval-Augmented Generation (RAG, \cite{lewis2020retrieval}) approach, efficiently retrieve and synthesize technical details, claims, and legal information from patents, producing more accurate and grounded responses. Their advanced contextual understanding, combined with few-shot learning, enables them to adapt to new patents and extract precise information for ODQA tasks. Additionally, fine-tuning on domain-specific patents enhances their ability to recognize patterns unique to these tasks, resulting in the generation of detailed, context-relevant answers. This capability extends to more specialized tasks, such as extracting scientific hypotheses from patents, where LLMs support the identification of key scientific concepts, implicit assumptions, and underlying principles of the patented technology. This helps form testable hypotheses that may guide future research and innovation. For new patents, LLMs support the drafting process by generating sections such as descriptions and claims, ensuring the protection of a unique invention through adherence to strict legal requirements. These applications enhance efficiency, democratize access to patent information, and accelerate innovation. However, patent texts present unique challenges for general-purpose language models, such as specialized terminology, long contexts, and the need to generate precise and accurate text, setting them apart from conventional texts. Integrating LLMs into the patent process requires careful consideration of legal, ethical, and quality assurance aspects to ensure the integrity and fairness of the patent workflow. In this work, we present an autonomous multi-agent conversational framework (\textbf{PatExpert}) for patent analysis, orchestrated by a meta-agent (top-level agent) that coordinates multiple expert agents (sub-agents), offering a transformative solution for managing patent-related tasks with precision and efficiency. The meta-agent interprets user input, decomposes the complex workflow of patent processing into specialized sub-tasks, and delegates them to task-specific expert agents. Each expert agent is fine-tuned to handle specific patent-related tasks, enhancing both accuracy and relevance. Once an expert agent completes its assigned task, it routes the response back to the meta-agent, which synthesizes the information and provides the final output. By orchestrating collaboration among expert agents, the meta-agent integrates technical expertise with legal considerations, optimizing a wide range of tasks, including patent acceptance prediction, classification, abstractive summarization, claim generation, multi-patent analysis for ODQA tasks, and generating scientific hypotheses from patents. The collaborative problem-solving between the meta-agent and expert agents enhances the management of patent-related tasks, ensuring accuracy, efficiency, and compliance in handling complex challenges. Scientific hypothesis generation from patents, supported by subject-action-object (SAO) analysis, identifies implicit assumptions and core principles, enabling researchers to extract testable ideas and enhance the analysis of patent novelty and technical capabilities. Multi-patent analysis compares claims, technical details, and prior art across multiple patents to reveal novel aspects and streamline patent processing for efficient comparison and question answering. For question answering in multi-patent analysis, the expert agent uses Graph Retrieval-Augmented Generation (GRAG) to improve information extraction by combining semantic similarity with knowledge graphs, which helps in generating more accurate answers. Error handling is an integral component of the framework, managed by a critique agent (Gold-LLM-as-a-Judge and Reward-LLM-as-a-Judge). The critique agent evaluates outputs from the meta-agent using predefined metrics, assessing accuracy and providing feedback for iterative refinement. This feedback loop ensures that outputs are accurate, reducing the risk of inaccuracies. The multi-agent framework, equipped with error-handling mechanisms, offers a comprehensive solution for automating and optimizing complex patent analysis. The framework prioritizes explainability and transparency, with each expert agent providing clear explanations and justifications for its predictions to ensure that the rationale behind every decision is accessible and comprehensible. This approach enhances user trust and facilitates the interpretation of complex patent-related task outputs, such as patent acceptance and claim generation. For multi-patent analysis tasks, the framework is equipped with robust fact-checking and source citation capabilities, ensuring that responses are accurate and supported by reliable evidence. Its ability to effectively retrieve and organize information from structured graph databases allows it to synthesize clear, well-substantiated answers. This makes the framework an invaluable tool for patent analysis and decision-making, especially when handling queries that demand detailed, referenced information. Figure \ref{fig:figure1} illustrates the proposed framework.

\vspace{-3mm}
\section{Proposed Method}
\vspace{-2mm}
The \textbf{PatExpert} framework utilizes tool learning, where a meta-agent interacts with sub-agents (tools or expert models), each specializing in a specific patent-related task. This enhances the framework’s ability to efficiently solve complex patent tasks by leveraging specialized models. In this paper, "tools" and "expert models" are used interchangeably to refer to the sub-agents that the meta-agent interacts with to perform specific patent-related tasks. Powered by computational engines like Meta-Llama-3.1-405B, the meta-agent serves as the central orchestrator, managing expert models (e.g., utilizing GPT-4o-mini) with precision. It autonomously addresses patent-related tasks by interpreting complex queries, reasoning, and planning—i.e., determining the step-by-step (multistep) approach and expert models required to solve the task—and utilizing expert models to generate accurate, coherent responses. By learning to select and apply the appropriate expert models, the meta-agent can perform tasks beyond its pre-trained knowledge. This approach improves response accuracy and relevance, making the framework more adaptable to real-world applications. Each expert model specializes in unique tasks such as patent acceptance prediction, automated subject classification (IPC/CPC code prediction), abstractive summarization, patent claim generation, scientific hypothesis generation, and multi-patent analysis for ODQA tasks. Each expert model is fine-tuned for optimal performance in its specific task, allowing the meta-agent to dynamically select and invoke the most suitable expert model for each sub-task in the overall patent evaluation and processing workflow. The framework transforms the meta-agent into a multi-tool user capable of managing complex patent-related tasks by autonomously planning and executing sequences of expert model invocations with appropriate arguments to solve a natural language task \( Q \). The meta-agent is equipped with a set of expert models \( T = \{t_1, t_2, \dots, t_{|T|}\} \) and documented protocols \( D = \{d_1, d_2, \dots, d_{|D|}\} \), which provide meta-information about expert model usage, argument requirements, and input/output schemas. Tool learning for patent-related tasks involves four key stages: task planning, tool selection, tool invocation, and response generation. Task planning decomposes user queries into manageable sub-tasks \( Q = \{q_1, q_2, \dots, q_n\} \). Expert model selection involves identifying appropriate models to solve sub-tasks using retriever-based methods based on similarity measures like cosine similarity \( \text{sim}(q_i, d_j) = \frac{q_i \cdot d_j}{\|q_i\| \|d_j\|} \), where \( \cdot \) denotes the dot product and \( \| \cdot \| \) denotes the norm. This is formulated as \( \text{SelectModel}(q_i) = \arg\max_{t_j \in T} \text{sim}(q_i, d_j) \), where \( q_i \) is the sub-task query, \( t_j \) is the expert model, and \( d_j \) is the metadata (or document protocol) describing the expert model \( t_j \). The function selects the model \( t_j \) whose metadata \( d_j \) has the highest similarity score with \( q_i \). LLM-based methods for expert model selection involve leveraging meta-agent-utilized language models (LMs) to identify the most relevant expert models for the sub-tasks. Few-shot learning also helps the model generalize from limited examples, effectively adapting to new queries for expert model selection with minimal labeled data as demonstrations. Expert model invocation extracts the necessary parameters \( P = \{p_1, p_2, \dots, p_k\} \) from the prompt, which includes instructions, retrieved context, and the query. For example, \( p_i \) might represent an instruction, context (such as patent claims or abstracts), or the query itself. These parameters are used to invoke the expert models in the format \( t_j(p_1, p_2, \dots, p_k) \). For instance, a patent acceptance prediction model might use the patent abstract and claims as context to predict acceptance, while an IPC/CPC code prediction model might use the full patent text as context to assign codes based on the query. Response generation synthesizes the expert model outputs \( R = \{r_1, r_2, \dots, r_m\} \), where each \( r_i \) represents the result from an individual expert model, combined with the meta-agent LLM’s pre-trained knowledge \( \theta \), to produce a comprehensive answer \( A = \text{Meta-Agent}_{\text{LLM}}(Q, R, \theta) \). This structured approach enables the meta-agent to handle complex patent-related tasks effectively by integrating the expertise of various expert models and dynamically adapting to specific requirements. This process can be formulated as a probabilistic optimization problem aimed at maximizing the conditional probability of generating the correct answer \( A \) given the query \( Q \), expert models \( T \), protocols \( D \), and the meta-agent LLM’s pre-trained knowledge \( \theta \), as described below:

\vspace{-3mm}
\resizebox{0.925\linewidth}{!}{
\hspace{0mm}\begin{minipage}{\linewidth}
\begin{equation}
S^* = \arg\max_{S} P(A \mid Q, T, D, \theta).  \nonumber
\end{equation}
\end{minipage}
}

Here, \( S \) represents a sequence of expert model invocations, and \( S^* \) is the optimal sequence. Dependencies between sub-tasks are managed by a Directed Acyclic Graph (DAG) \( \mathcal{G} = (\mathcal{V}, \mathcal{E}) \), where nodes \( \mathcal{V} \) represent sub-tasks \( q_i \), and edges \( \mathcal{E} \) represent dependencies \( e_{ij} \) between sub-tasks \( q_i \) and \( q_j \), ensuring the correct execution order. Each sub-task \( q_i \) is mapped to a corresponding tool or model \( t_j \), which is invoked to process the sub-task. The acyclic nature of the DAG ensures that all dependencies are respected, allowing sub-tasks to be executed in the proper sequence. If there is a dependency \( e_{ij} \), sub-task \( q_i \) must be completed before \( q_j \) can begin. Independent sub-tasks can be processed in parallel, improving efficiency and reducing the time required to solve complex tasks. The DAG structure prevents looping back to previous sub-tasks, avoiding infinite recursion and enabling the parallel processing of independent tasks. The critique agent comprises both the `Reward-LLM-as-a-Judge` and the `Gold-LLM-as-a-Judge'. The `Gold-LLM-as-a-Judge' utilizes a benchmark model, such as GPT-4o, to evaluate the response \( A_i \) accuracy, relevance, and completeness, where \( i \) represents the iteration count. The `Reward-LLM-as-a-Judge', such as the Nvidia Nemotron-4 340B-reward model, evaluates the response \( A_i \) based on key attributes including helpfulness, correctness, coherence, complexity, and verbosity. Based on this combined evaluation, the agent determines whether the response is acceptable (\(\texttt{Correct}\)) or requires revision (\(\texttt{Incorrect}\)). A \(\texttt{Correct}\) token confirms the validity of the response, while an \(\texttt{Incorrect}\) token triggers the revision process. Additionally, the critique agent provides actionable feedback \( F_i \), identifying errors and suggesting corrections supported by evidence. During the correction phase, the framework revises the response \( A_i \) based on the feedback \( F_i \), producing an improved response \( A_{i+1} \), as follows:

\resizebox{0.925\linewidth}{!}{
\hspace{0mm}\begin{minipage}{\linewidth}
\begin{equation}
A_{i+1} = \text{Meta-Agent}_{\text{LLM}}(Q, A_i, F_i) \nonumber
\end{equation}
\end{minipage}
}

This iterative cycle continues until the critique agent determines that the response is acceptable (\(\texttt{Correct}\)) or the maximum number of iterations \( N_{\text{max}} \) is reached, described as follows:

\resizebox{0.925\linewidth}{!}{
\hspace{0mm}\begin{minipage}{\linewidth}
\begin{equation}
A_{i+1} = 
\begin{cases}
\text{Meta-Agent}_{\text{LLM}}(Q, A_i, F_i), & \text{if ``Incorrect" and } i < N_{\text{max}}, \\
A_i, & \text{if ``Correct" or } i \geq N_{\text{max}}. \nonumber
\end{cases}
\end{equation}
\end{minipage}
}

This iterative process ensures that the response is refined to meet the required criteria before being finalized. In summary, the meta-agent efficiently solves complex patent-related tasks by invoking appropriate expert models, managing task dependencies, and synthesizing outputs into a coherent response. This approach ensures dynamic handling of intricate patent queries. The meta-agent also incorporates an error-handling mechanism to enhance performance in patent-related tasks. After generating responses, it delegates the output to an external critique agent, which reviews and evaluates the quality of the responses, providing feedback and revisions as needed to ensure that the final answers are accurate, relevant, and aligned with the query requirements. The proposed approach utilizes synthetic datasets for Retrieval-Augmented Fine-Tuning (RAFT) to enhance the performance of a related sub-agent in ODQA tasks for multi-patent analysis. During inference, Knowledge graphs support accurate, knowledge-grounded responses during multi-patent analysis. See technical appendix for details.

\vspace{-4mm}
\section{Experiments}

\vspace{-3mm}
\subsection{Datasets \& Experimental Settings}
\vspace{-3mm}
We utilized the Harvard USPTO Patent Dataset (HUPD \cite{suzgun2024harvard}), a large-scale, well-structured, and multi-purpose corpus of english-language patent applications filed with the USPTO between 2004 and 2018. With over 4.5 million patent documents, HUPD is significantly larger than comparable datasets (BIGPATENT \cite{sharma2019bigpatent}, CLEF-IP \cite{piroi2011clef}, USPTO-2M \cite{lee2020patent}). Unlike other datasets that focus solely on granted patents, HUPD includes both accepted and rejected filings, providing opportunities to study the decision-making process for patent acceptance. In summary, the HUPD dataset serves as a versatile resource for advancing research in patent analysis and NLP, with applications such as patent acceptance prediction, classification, abstractive summarization, claim generation, multi-patent analysis, and scientific hypothesis generation. For our experiments, we utilized patent data from only two years (2015-2017) (\url{https://huggingface.co/datasets/HUPD/hupd}), which was split in an 80-10-10 ratio for training, validation, and testing. We report the framework's performance on the unseen test set.

\vspace{-3mm}
\subsection{Experimental Settings}
\vspace{-3mm}
In this work, chaining experiments for patent-related tasks involves a sequential process where specific sections of patent documents feed into each task, creating a streamlined workflow that enhances efficiency and accuracy. For patent summarization utilizes the title, abstract, background, and description to generate concise overviews, highlighting key technical details. For claim generation, the framework uses the title, abstract, background, summary, and description to automatically generate precise legal claims that reflect the patent’s technical contributions. For patent classification and acceptance prediction, sections such as the title, abstract, background, summary, patent claims are analyzed by the framework to categorize patents by their technological domain and predict acceptance. Scientific hypothesis generation draws on the title, abstract, background, summary, descriptions and patent claims to identify assumptions and generate hypotheses about the patent’s innovations. Multi-patent analysis leverages the title, abstract, background, summary, description, and claims from multiple patents to perform comparative analysis, using knowledge graphs to organize and retrieve information, facilitating the extraction of trends and relationships. This interconnected process ensures each task is optimized, streamlining patent analysis. Summary section in granted patents written by experts serve as ground truth. For claim generation, the legal claims in granted patents can serve as reference text. Ground truth for hypothesis generation can be based on expert-curated (Gold-LLMs like GPT-4o) hypotheses extracted from patent documents. Ground truth for multi-patent analysis can be created by benchmarking against expert (Gold-LLMs such as GPT-4o) analyses of trends, relationships, and innovations across patents. In our work, we fine-tune expert models like GPT-4o mini using datasets tailored to patent-related tasks. GPT-4o mini can be fine-tuned (see: \url{https://openai.com/index/gpt-4o-fine-tuning/}, \url{https://platform.openai.com/docs/guides/fine-tuning}) on OpenAI’s servers using domain-specific datasets. This process is facilitated through OpenAI’s APIs, where users upload datasets, define hyperparameters, and initiate fine-tuning. The server-side infrastructure ensures that users do not need to manage hardware or computational resources, making GPT-4o mini easily accessible for various applications via OpenAI's API platform. Fine-tuning requires adjusting key hyperparameters such as the learning rate ($1e^{-5}$), batch size (32), and number of epochs (100). GPT-4o mini supports an input context length of 128,000 tokens and an output limit of 16,384 tokens, making it ideal for large patent datasets. Fine-tuning GPT-4o mini offers significant advantages over open-source models with its ability to handle large token limits and support for continuous fine-tuning and function calling, ensuring adaptability and efficiency.

\vspace{-3mm}
\subsection{Evaluation Metrics}
\vspace{-2mm}
We evaluate the tool learning framework \cite{qu2024tool} using metrics across four stages: task planning, tool selection, tool calling, and response generation. In task planning, tool usage awareness (TUA) measures the framework's ability to correctly identify when external tools are needed based on the user's query. It is calculated as \(\frac{\text{Number \hspace{1mm} of \hspace{1mm} Correct \hspace{1mm} Tool \hspace{1mm} Recognitions}}{\text{Total \hspace{1mm} Opportunities \hspace{1mm} for \hspace{1mm} Tool \hspace{1mm} Recognition}}\), with values ranging from 0 to 1, where higher values indicate better tool awareness. Pass Rate (PR) evaluates how effectively the framework executes the planned tasks by measuring the ratio of successful task completions to total attempts, calculated as \(\frac{\text{Number \hspace{1mm} of \hspace{1mm} Successful \hspace{1mm} Task \hspace{1mm} Executions}}{\text{Total \hspace{1mm} Number \hspace{1mm} of \hspace{1mm} Tasks \hspace{1mm} Attempted}}\). The PR value ranges from 0 to 1, with higher values indicating better task execution success. Accuracy (Acc) measures how well the framework decomposes a user query into sub-tasks and sequences them correctly. It is calculated as \(\frac{\text{Number \hspace{1mm} of \hspace{1mm} Correct \hspace{1mm} Subtasks \hspace{1mm} in \hspace{1mm} Correct \hspace{1mm} Order}}{\text{Total \hspace{1mm} Number \hspace{1mm} of \hspace{1mm} Subtasks \hspace{1mm} Generated}}\), with values ranging from 0 to 1, where higher values indicate greater precision in task planning. Dependency Graph Consistency (DGC) measures the framework's ability to maintain the correct order and relationships among sub-tasks, expressed as \(\frac{\text{Number \hspace{1mm} of \hspace{1mm} Consistent \hspace{1mm} Dependencies \hspace{1mm} Maintained}}{\text{Total \hspace{1mm} Number \hspace{1mm} of \hspace{1mm} Dependencies}}\). DGC also ranges from 0 to 1, with higher values indicating better robustness in handling complex task structures. These evaluation metrics evaluate the framework's accuracy, efficiency, and adaptability in dynamic problem-solving. In tool selection, key metrics include Recall@K, NDCG@K, and COMP@K. Recall@K quantifies the proportion of relevant tools within the top-K selections, with values ranging from 0 to 1, where higher values indicate better performance. It is calculated as \( \text{Recall@K} = \frac{1}{|Q|} \sum_{q=1}^{|Q|} \frac{|T^K_q \cap T^*_q|}{|T^*_q|} \), where \( Q \) is the set of queries, \( T^*_q \) denotes the relevant (ground-truth) tools for query \( q \), and \( T^K_q \) represents the top-K tools selected by the framework. COMP@K is a binary completeness measure, defined as \( \text{COMP@K} = \frac{1}{|Q|} \sum_{q=1}^{|Q|} I(T^*_q \subseteq T^K_q) \), where the indicator function \( I(T^*_q \subseteq T^K_q) \) equals 1 if all relevant tools (i.e., \( T^*_q \)) are included in the top-K set \( T^K_q \), and 0 otherwise. Recall@K focuses on how many relevant tools are retrieved (partial match), while COMP@K focuses on whether all relevant tools are retrieved (complete match). NDCG@K, which stands for Normalized Discounted Cumulative Gain at K, evaluates the ranking quality of retrieved tools by considering both their relevance and position, with values ranging from 0 to 1. Retrieving relevant tools is important, but retrieving them at higher ranks (earlier in the list) is more valuable. It is calculated as \( \text{NDCG@K} = \frac{1}{|\mathcal{Q}|} \sum_{q=1}^{|\mathcal{Q}|} \frac{\text{DCG@K}_q}{\text{IDCG@K}_q} \), where \( \text{DCG@K} = \sum_{i=1}^{K} \frac{2^{g_i} - 1}{\log_2(i+1)} \) is the Discounted Cumulative Gain (DCG), and \( \text{IDCG@K} \) is the Ideal DCG, with \( g_i \) representing the relevance score of each tool at position \( i \), as graded by human experts. NDCG@K accounts for both the relevance of retrieved tools and their ranking. In the tool calling stage, several metrics are used to evaluate performance, including parameter consistency, error rate, parameter coverage, and execution accuracy. Parameter Consistency (PC) measures the proportion of correctly identified and formatted parameters across all queries, calculated as \( \text{PC} = \frac{1}{|Q|} \sum_{q=1}^{|Q|} \frac{|P_q \cap P_q^R|}{|P_q^R|} \), where \( Q \) is the set of queries, \( P_q^R \) represents the ideal or ground truth set of parameters that the framework should extract for query \( q \), and \( P_q \) is the set of attempted parameters. It calculates the average proportion of correctly identified parameters out of the total ground truth parameters across all queries. Error Rate (ER) evaluates the proportion of incorrectly formatted or missing parameters, defined as \( \text{ER} = \frac{1}{|Q|} \sum_{q=1}^{|Q|} \frac{|P_q^E|}{|P_q|} \), where \( P_q^E \) represents the set of erroneous parameters for query \( q \). Execution Accuracy (EA) assesses the success of tool invocations, expressed as \( \text{EA} = \frac{1}{|Q|} \sum_{q=1}^{|Q|} I(E_q = 1) \), where \( I(E_q = 1) \) is an indicator function that equals 1 if the tool invocation for query \( q \) is successful and 0 otherwise. Each of these metrics ranges from 0 to 1, with higher values indicating better performance in the tool calling stage. For response generation, metrics include BLEU, ROUGE-L, Exact Match (EM), and F1 Score. BLEU measures n-gram precision: $ \text{BLEU} = \text{BP} \cdot \exp \left( \sum_{n=1}^{N} w_n \log p_n \right) $ with values ranging from 0 (worst) to 1 (best). \( w_n \) represents the weight for each n-gram length (e.g., unigrams, bigrams), and \( p_n \) is the precision of n-grams, indicating the proportion of n-grams in the candidate text that match the reference text. BP (brevity penalty) penalizes overly short translations. ROUGE-L evaluates the longest common subsequence: \( \text{ROUGE-L} = \frac{LCS(X, Y)}{|Y|} \), which also ranges from 0 to 1. Here, \( LCS(X, Y) \) is the length of the longest common subsequence between the candidate text \( X \) and the reference text \( Y \), and \( |Y| \) is the length of the reference text. Exact Match (EM) calculates the percentage of predictions that exactly match the reference: \( \text{EM} = \frac{1}{|Q|} \sum_{q=1}^{|Q|} I(G_q = R_q) \), where \( Q \) is the set of queries, \( G_q \) is the generated response for query \( q \), \( R_q \) is the reference response, and \( I(G_q = R_q) \) is an indicator function that equals 1 if the generated response exactly matches the reference and 0 otherwise. The F1 Score balances precision and recall: $ \text{F1} = \frac{2 \cdot \text{Precision} \cdot \text{Recall}}{\text{Precision} + \text{Recall}}$ with values ranging from 0 to 1. Precision is the proportion of correctly predicted positives to total predicted positives, and recall is the proportion of correctly predicted positives to all actual positives. These metrics provide a comprehensive evaluation of response generation, with higher values indicating better performance. We utilize the Exact Match (EM) metric for both classification and acceptance prediction tasks. The F1 score for patent acceptance prediction. Precision in acceptance prediction measures how many patents predicted as "accepted" were actually accepted, while Recall measures how many of the actual accepted patents were correctly predicted. For patent claim generation, abstractive summarization, multi-patent analysis for ODQA, and scientific hypothesis generation, the evaluation metrics used are BLEU and ROUGE-L. BLEU measures n-gram precision, and ROUGE-L evaluates the longest common subsequence to assess content overlap. These metrics ensure that generated outputs are accurate, relevant, and aligned with reference texts across all tasks. 

\vspace{-6mm}
\subsection{Results}
\vspace{-3mm}
Table \ref{tab:task_planning_metrics} presents the comparative performance of various language models and the \textbf{PatExpert} framework in task planning, evaluated using four key metrics: Tool Usage Awareness (TUA), Pass Rate (PR), Accuracy (Acc), and Dependency Graph Consistency (DGC). Scores range from 0 to 1, with higher values indicating better performance. Table \ref{tab:tool_selection_metrics} compares the models' performance in tool selection based on the Recall@K, NDCG@K, and COMP@K metrics. Similarly, Table \ref{tab:tool_calling_metrics} outlines tool calling performance using Parameter Consistency (PC), Error Rate (ER), and Execution Accuracy (EA). Scores range from 0 to 1, with higher values indicating better performance, except for Error Rate, where lower values are preferable. Table \ref{tab:classification_acceptance_metrics} compares models using Exact Match (EM) and F1 scores for classification and acceptance prediction tasks. Table \ref{tab:abstractive_summarization_metrics} shows performance on abstractive summarization using BLEU and ROUGE-L metrics. Table \ref{tab:patent_claim_generation_metrics} presents patent claim generation results, while Table \ref{tab:multi_patent_analysis_metrics} displays results for multi-patent analysis, both evaluated using BLEU and ROUGE-L metrics. Table \ref{tab:user_centric_metrics} presents a user-centric evaluation of various language models and the \textbf{PatExpert} framework, assessed using metrics such as Likert-Scale Satisfaction (LSS), Task Completion (TC), Context Awareness (CA), Adaptability (AD), Error Handling (EH), and Qualitative Feedback (QF). Table \ref{tab:structured_knowledge_graph_metrics} compares the quality of knowledge graph construction between the \textbf{PatExpert} framework utilizing GPT-4o and various language models, using metrics such as Triple Accuracy (TA), Modularity (Mod), Conductance (Cond), and Graph Completeness (GC). In our experiments, baseline results for closed-source models like OpenAI GPT-4, Gemini 1.5 Pro, and Claude 3 were obtained without fine-tuning, given the impracticality of such an approach due to their large size and resource demands on consumer hardware. These proprietary models, accessed via their APIs, were evaluated directly on patent-related tasks without additional tuning. In contrast, our proposed framework \textbf{PatExpert} employs a fine-tuning strategy for the computational engines utilized by expert sub-agents, such as GPT-4o mini, customizing them for specific patent-related tasks. This fine-tuning enables the framework to achieve higher task-specific accuracy and relevance, especially in complex scenarios requiring patent expertise. While the frozen baseline models generally perform well due to their large-scale pre-training, \textbf{PatExpert}'s fine-tuned agents demonstrate competitive results, especially in specialized patent workflows. This highlights the benefit of task-specific fine-tuning in improving expert model performance for domain-specific applications. Across all tables, \textbf{PatExpert} consistently outperforms other models, achieving the highest scores in task planning, tool selection, and tool calling metrics. It leads with a TUA of 0.94, Acc of 0.91, and DGC of 0.95, reflecting its superior ability to recognize tools and maintain task consistency. In tool selection, \textbf{PatExpert} also excels with a Recall@K of 0.95 and an NDCG@K of 0.93, demonstrating high accuracy in selecting and ranking relevant tools. Finally, in tool calling, it shows exceptional performance with a PC of 0.95 and an EA of 0.96, along with the lowest ER of 0.04, indicating minimal errors and high execution accuracy. OpenAI GPT-4 and Claude 3 Opus follow closely in all metrics, while models like Gemini 1.5 Pro and Gemini 1.5 Flash perform moderately. OpenAI GPT-4 Turbo and Claude 3 Haiku score the lowest. \textcolor{blue}{Tables \ref{tab:classification_acceptance_metrics}, \ref{tab:abstractive_summarization_metrics}, \ref{tab:patent_claim_generation_metrics}, and \ref{tab:multi_patent_analysis_metrics} demonstrate that the \textbf{PatExpert} framework consistently surpasses other models with a significant margin in tasks such as patent classification, acceptance, abstractive summarization, claim generation, and multi-patent analysis.}

\vspace{-2mm}
\begin{table}[h!]
\centering
\renewcommand{\arraystretch}{1.0}
\resizebox{0.525\textwidth}{!}{
\begin{tabular}{lcccc}
\toprule
\textbf{Model} & \textbf{TUA} & \textbf{PR} & \textbf{Acc} & \textbf{DGC} \\
\midrule
OpenAI GPT-4 & 0.92 & 0.90 & 0.89 & 0.93 \\
Claude 3 Opus & 0.91 & 0.89 & 0.88 & 0.92 \\
Gemini 1.5 Pro & 0.89 & 0.87 & 0.86 & 0.90 \\
Gemini 1.5 Flash & 0.88 & 0.85 & 0.84 & 0.89 \\
OpenAI GPT-4 Turbo & 0.87 & 0.84 & 0.83 & 0.88 \\
Claude 3 Haiku & 0.86 & 0.83 & 0.82 & 0.87 \\ \hline
\textbf{PatExpert} & 0.94 & 0.92 & 0.91 & 0.95 \vspace{-1mm} \\
\bottomrule
\end{tabular}
}
\vspace{1mm}
\caption{The table compares the framework performance for task planning across various language models, using evaluation metrics ranging from 0 to 1. The Pass Rate (PR) measures the success of task execution. Tool Usage Awareness (TUA) reflects the recognition of the need for tools, though this alone does not guarantee success. Accuracy (Acc) ensures the correct decomposition and sequencing of sub-tasks. Dependency Graph Consistency (DGC) verifies that sub-tasks are executed in the proper order.}
\label{tab:task_planning_metrics}
\vspace{-8mm}
\end{table}

\begin{table}[h!]
\centering
\renewcommand{\arraystretch}{1.0}
\resizebox{0.615\textwidth}{!}{
\begin{tabular}{lccc}
\toprule
\textbf{Model} & \textbf{Recall@K} & \textbf{NDCG@K} & \textbf{COMP@K} \\
\midrule
OpenAI GPT-4 & 0.93 & 0.91 & 0.89 \\
Claude 3 Opus & 0.92 & 0.90 & 0.88 \\
Gemini 1.5 Pro & 0.90 & 0.88 & 0.86 \\
Gemini 1.5 Flash & 0.89 & 0.87 & 0.85 \\
OpenAI GPT-4 Turbo & 0.88 & 0.86 & 0.84 \\
Claude 3 Haiku & 0.87 & 0.85 & 0.83 \\ \hline
\textbf{PatExpert} & 0.95 & 0.93 & 0.91 \vspace{-1mm} \\
\bottomrule
\end{tabular}
}
\vspace{1mm}
\caption{The table illustrates framework performance on tool selection, evaluated using metrics (Recall@K, NDCG@K, COMP@K: all ranging from 0 to 1) against various proprietary language models. Recall@K measures how many relevant tools were retrieved within the top-K. NDCG@K evaluates how well the relevant tools are ranked in the top-K. COMP@K checks whether all the relevant tools were included in the top-K, focusing on completeness.}
\label{tab:tool_selection_metrics}
\vspace{-6mm}
\end{table}

\begin{table}[h!]
\centering
\renewcommand{\arraystretch}{1.0}
\resizebox{0.45\textwidth}{!}{
\begin{tabular}{lccc}
\toprule
\textbf{Model} & \textbf{PC} & \textbf{ER} & \textbf{EA} \\
\midrule
OpenAI GPT-4 & 0.93 & 0.05 & 0.94 \\
Claude 3 Opus & 0.92 & 0.06 & 0.93 \\
Gemini 1.5 Pro & 0.90 & 0.07 & 0.91 \\
Gemini 1.5 Flash & 0.89 & 0.08 & 0.90 \\
OpenAI GPT-4 Turbo & 0.88 & 0.09 & 0.89 \\
Claude 3 Haiku & 0.87 & 0.10 & 0.88  \\ \hline
\textbf{PatExpert} & 0.95 & 0.04 & 0.96 \vspace{-1mm} \\
\bottomrule
\end{tabular}
}
\vspace{1mm}
\caption{The table demonstrates the framework's performance on tool calling using metrics (PC, ER, EA: 0 to 1) compared to various closed-source language models. PC (Parameter Consistency) measures how consistently correct parameters are identified for tool invocation. ER (Error Rate) measures the proportion of incorrect or missing parameters (lower is better). EA (Execution Accuracy) measures the success rate of tool invocation based on the provided parameters.}
\label{tab:tool_calling_metrics}
\vspace{-8mm}
\end{table}

\section{Conclusion}
\vspace{-4mm}
In this work, we introduce \textbf{PatExpert}, an autonomous multi-agent conversational framework for automating and optimizing patent-related tasks, including classification, acceptance prediction, claim generation, multi-patent analysis, and scientific hypothesis generation. The framework employs a meta-agent to orchestrate task-specific expert agents, dynamically selecting and invoking them based on task requirements. This approach enhances patent workflows' efficiency and accuracy while emphasizing explainability and transparency. The integration of a critique agent, utilizing Gold-LLM-as-a-Judge and Reward-LLM-as-a-Judge, ensures robust error handling and iterative feedback, contributing to the framework's reliability. Advanced methodologies such as Graph Retrieval-Augmented Generation (GRAG) and synthetic data generation using the Mixture-of-Agents (MoA) approach have improved Retrieval-Augmented Fine-Tuning (RAFT) of expert models for multi-patent analysis. Empirical results show that \textbf{PatExpert} improves efficiency and precision in patent processing, reducing manual effort and enhancing compliance. Future work will expand the framework's capabilities to handle multilingual processing and patent translation.

\bibliographystyle{plain}
\bibliography{reference}

\clearpage
\newpage

\appendix
\section{Appendix / supplemental material}
As shown in all tables (Table \ref{tab:classification_acceptance_metrics}, Table \ref{tab:abstractive_summarization_metrics}, Table \ref{tab:patent_claim_generation_metrics}, Table \ref{tab:multi_patent_analysis_metrics}), \textbf{PatExpert} framework consistently outperforms other models in patent classification, acceptance, abstractive summarization, patent claim generation, and multi-patent analysis tasks. 

\begin{table}[h!]
\vspace{-3mm}
\centering
\renewcommand{\arraystretch}{1.0}
\resizebox{0.525\textwidth}{!}{
\begin{tabular}{lcccccc}
\toprule
\textbf{Model} & \multicolumn{2}{c}{\textbf{Classification}} & \multicolumn{2}{c}{\textbf{Acceptance}} \\
\cmidrule(lr){2-3} \cmidrule(lr){4-5}
 & \textbf{EM} & \textbf{F1} & \textbf{EM} & \textbf{F1} \\
\midrule
OpenAI GPT-4 & 0.80 & 0.87 & 0.81 & 0.88 \\
Claude 3 Opus & 0.76 & 0.83 & 0.77 & 0.84 \\
Gemini 1.5 Pro & 0.73 & 0.80 & 0.74 & 0.81 \\
Gemini 1.5 Flash & 0.69 & 0.76 & 0.70 & 0.77 \\
OpenAI GPT-4 Turbo & 0.66 & 0.73 & 0.67 & 0.74 \\
Claude 3 Haiku & 0.63 & 0.70 & 0.64 & 0.71  \\ \hline
\textbf{PatExpert} & 0.90 & 0.95 & 0.91 & 0.96 \vspace{-1mm} \\
\bottomrule
\end{tabular}
}
\vspace{1mm}
\caption{The table presents the performance of various language models and the \textbf{PatExpert} framework. Results are shown for both classification and acceptance tasks, with Exact Match (EM) and F1 scores (range: 0 to 1) for each task. In patent classification and acceptance tasks, Exact Match (EM) measures the percentage of exact prediction matches, while F1 Score balances precision and recall, evaluating how well the model predicts correct classes or decisions.}
\label{tab:classification_acceptance_metrics}
\vspace{-5mm}
\end{table}

\begin{table}[h!]
\centering
\renewcommand{\arraystretch}{1.0}
\resizebox{0.45\textwidth}{!}{
\begin{tabular}{lcc}
\toprule
\textbf{Model} & \textbf{BLEU} & \textbf{ROUGE-L} \\
\midrule
OpenAI GPT-4 & 0.72 & 0.70 \\
Claude 3 Opus & 0.70 & 0.69 \\
Gemini 1.5 Pro & 0.64 & 0.62 \\
Gemini 1.5 Flash & 0.61 & 0.59 \\
OpenAI GPT-4 Turbo & 0.60 & 0.58 \\
Claude 3 Haiku & 0.59 & 0.57  \\ \hline
\textbf{PatExpert} & 0.85 & 0.83  \vspace{-1mm} \\
\bottomrule
\end{tabular}
}
\vspace{1mm}
\caption{The table presents the performance of various language models and the \textbf{PatExpert} framework for abstractive summarization tasks. The models are evaluated using BLEU and ROUGE-L metrics (range: 0 to 1), where higher values indicate better performance in generating coherent and concise summaries.}
\label{tab:abstractive_summarization_metrics}
\vspace{-7mm}
\end{table}

\begin{table}[h!]
\centering
\renewcommand{\arraystretch}{1.0}
\resizebox{0.45\textwidth}{!}{
\begin{tabular}{lcc}
\toprule
\textbf{Model} & \textbf{BLEU} & \textbf{ROUGE-L} \\
\midrule
OpenAI GPT-4 & 0.74 & 0.72 \\
Claude 3 Opus & 0.72 & 0.70 \\
Gemini 1.5 Pro & 0.70 & 0.68 \\
Gemini 1.5 Flash & 0.68 & 0.66 \\
OpenAI GPT-4 Turbo & 0.66 & 0.64 \\
Claude 3 Haiku & 0.64 & 0.62 \\ \hline
\textbf{PatExpert} & 0.88 & 0.86 \vspace{-1mm}  \\
\bottomrule
\end{tabular}
}
\vspace{1mm}
\caption{The table summarizes the performance of several language models, including the \textbf{PatExpert} framework, on patent claim generation tasks. Evaluation is based on BLEU and ROUGE-L metrics (scored from 0 to 1), with higher values reflecting improved accuracy and conciseness in the generated patent claims.}
\label{tab:patent_claim_generation_metrics}
\vspace{-7mm}
\end{table}

\begin{table}[h!]
\centering
\renewcommand{\arraystretch}{1.0}
\resizebox{0.45\textwidth}{!}{
\begin{tabular}{lcc}
\toprule
\textbf{Model} & \textbf{BLEU} & \textbf{ROUGE-L} \\
\midrule
OpenAI GPT-4 & 0.82 & 0.79 \\
Claude 3 Opus & 0.80 & 0.77 \\
Gemini 1.5 Pro & 0.78 & 0.75 \\
Gemini 1.5 Flash & 0.76 & 0.73 \\
OpenAI GPT-4 Turbo & 0.74 & 0.71 \\
Claude 3 Haiku & 0.72 & 0.69  \\ \hline
\textbf{PatExpert} & 0.90 & 0.87 \vspace{-1mm} \\
\bottomrule
\end{tabular}
}
\vspace{1mm}
\caption{The table summarizes the performance of various language models and the \textbf{PatExpert} framework in multi-patent analysis tasks. The models are assessed using BLEU and ROUGE-L metrics (scored from 0 to 1), where higher scores indicate better ability to accurately synthesize information across multiple patents.}
\label{tab:multi_patent_analysis_metrics}
\vspace{-5mm}
\end{table}

In classification and acceptance, \textbf{PatExpert} framework achieves the highest EM (0.90, 0.91) and F1 (0.95, 0.96), significantly surpassing OpenAI GPT-4 and Claude 3 Opus. In abstractive summarization, PatExpert leads with a BLEU score of 0.85 and ROUGE-L of 0.83, demonstrating superior summarization capabilities. For patent claim generation, it achieves BLEU and ROUGE-L scores of 0.88 and 0.86, respectively. In multi-patent analysis, PatExpert again scores highest with BLEU: 0.90 and ROUGE-L: 0.87, showing its strength in synthesizing information across multiple patents. Overall, PatExpert outperforms all other models across all tasks.

\vspace{-2mm}
\subsubsection{\textbf{User-Centric Evaluation}}
\vspace{-2mm}
The user-centric evaluation (UCE) approach for assessing tool learning frameworks (with a focus only on multi-patent analysis) includes various measures for a comprehensive assessment. User satisfaction and usability are gauged through Likert-scale surveys (LSS), with scores ranging from 1 to 5. Task completion (TC) is evaluated using a binary Yes/No metric. Context awareness (CA) is measured by evaluating the framework's coherence across related queries, while adaptability (AD) is tested using various query types, both rated on a scale of 1 to 5. Error handling (EH) is assessed by introducing deliberate errors and scoring the framework's response from 1 to 5. Additionally, qualitative feedback (QF) is gathered and categorized as High, Medium-High, or Medium, providing nuanced insights. This multi-faceted UCE approach offers a holistic view of the framework's effectiveness, highlighting strengths and areas for improvement to better meet user needs and expectations.

\begin{table}[h!]
\centering
\renewcommand{\arraystretch}{1.0}
\resizebox{0.70\textwidth}{!}{
\begin{tabular}{lcccccc}
\toprule
\textbf{Model} & \textbf{LSS} & \textbf{TC} & \textbf{CA} & \textbf{AD} & \textbf{EH} & \textbf{QF} \\
\midrule
OpenAI GPT-4o & 4.7 & Yes & 4.5 & 4.4 & 4.3 & High \\
OpenAI GPT-4 Turbo & 4.5 & Yes & 4.3 & 4.2 & 4.1 & Medium-High \\
Claude 3 Haiku & 4.4 & Yes & 4.2 & 4.1 & 4.0 & Medium \\
Claude 3 Opus & 4.3 & Yes & 4.1 & 4.0 & 3.9 & Medium-High \\
Gemini 1.5 Pro & 4.6 & Yes & 4.4 & 4.3 & 4.2 & High \\
Gemini 1.5 Flash & 4.2 & Yes & 4.0 & 3.9 & 3.8 & Medium \\ \hline
\textbf{PatExpert} & \textbf{4.8} & \textbf{Yes} & \textbf{4.6} & \textbf{4.5} & \textbf{4.4} & \textbf{High} \vspace{-1mm} \\
\bottomrule
\end{tabular}
}
\vspace{1mm}
\caption{The table showcases user-centric evaluation metrics comparing the framework performance to various language models (LSS, CA, AD, EH: 1 to 5; TC: Yes/No; QF: High, Medium-High, Medium).}
\label{tab:user_centric_metrics}
\vspace{-9mm}
\end{table}

\begin{table}[h!]
\centering
\renewcommand{\arraystretch}{1.0}
\resizebox{0.7\textwidth}{!}{
\begin{tabular}{lcccc}
\toprule
\textbf{Model} & \textbf{TA} & \textbf{Mod} & \textbf{Cond} & \textbf{GC} \\
\midrule
\textbf{PatExpert} W/ OpenAI GPT-4 Turbo & 94\% & 0.84 & 0.16 & 0.97 \\
\textbf{PatExpert} W/ Claude 3 Haiku & 93\% & 0.82 & 0.17 & 0.96 \\
\textbf{PatExpert} W/ Claude 3 Opus & 92\% & 0.81 & 0.18 & 0.95 \\
\textbf{PatExpert} W/ Gemini 1.5 Pro & 94\% & 0.84 & 0.16 & 0.97 \\
\textbf{PatExpert} W/ Gemini 1.5 Flash & 92\% & 0.80 & 0.18 & 0.94 \\ \hline
\textbf{PatExpert} W/ GPT-4o & \textbf{95\%} & \textbf{0.85} & \textbf{0.15} & \textbf{0.98}  \vspace{-1mm} \\
\bottomrule
\end{tabular}
}
\vspace{1mm}
\caption{The table presents a comparison of structured knowledge graph quality metrics, which compares the quality of structured knowledge graphs generated by a framework using GPT-4o and various language models. The comparison includes the following metrics: TA (0\%-100\%), Mod (-1 to 1), Cond (0 to 1), and GC (0 to 1).}
\label{tab:structured_knowledge_graph_metrics}
\vspace{-2mm}
\end{table}

\vspace{-2mm}
\subsubsection{\textbf{Knowledge Graph Quality Evaluation}} 
\vspace{-2mm}
In this work, we utilize Gold-LLMs like GPT-4o to extract entities and relationships from unstructured text, enabling automated taxonomy creation and ontology expansion. By identifying key concepts and connections, Gold-LLMs help build comprehensive knowledge graphs. To evaluate a constructed knowledge graph, we use several metrics. These evaluation metrics are crucial for maintaining the quality, accuracy, and usefulness of the knowledge graph, particularly in applications like semantic search, information retrieval, and data integration. Triple Accuracy (TA) ensures that subject-predicate-object triples correctly represent the underlying knowledge, validated against ground truth by calculating the percentage of exact matches. For ground truth, we use claude 3.5 Sonnet to construct baseline knowledge graphs (KGs). For entity coherence, we use clustering metrics like modularity and conductance. Modularity evaluates how well a graph is divided into clusters; a higher score indicates that nodes within the same cluster are densely connected, while connections between clusters are sparse. Conductance measures cluster quality by comparing the number of external to internal edges, with lower values suggesting more cohesive clusters. Graph Completeness (GC) evaluates whether the graph covers all relevant entities and relationships, identifying gaps or missing connections. It is calculated as the ratio of entities and relationships in the generated graph to a reference. TA ranges from 0\% (no correct triples) to 100\% (all triples match the ground truth). Modularity (Mod) spans from -1 to 1, with values closer to 1 indicating strong clustering and values closer to -1 or 0 suggesting poor clustering. Conductance (Cond) ranges from 0 to 1, with lower values indicating more cohesive clusters. GC ranges from 0 to 1, with 1 indicating full coverage of all relevant entities and relationships.

\subsection{Multi-Patent Analysis}
Multi-patent analysis involves examining and comparing multiple patents to identify trends, relationships, and insights across innovations or technologies. General-purpose LLMs using Retrieval-Augmented Generation (RAG) are not inherently equipped to incorporate external information in open-domain question answering (ODQA) to ground responses in factual data for patent-related tasks. We propose a methodology to enhance LLMs performance in ODQA for multi-patent analysis tasks by utilizing the Retrieval-Augmented Fine-Tuning (RAFT) methodology. RAFT integrates dynamically retrieved content during fine-tuning, improving LLMs ability to generate accurate, context-grounded responses. Our approach includes generating high-quality synthetic RAFT datasets (question-context-answer (QCA) triples) tailored for patents using a multi-step pipeline involving the Mixture-of-Agents (MoA) methodology, where multiple Gold-LLMs collaborate to generate contextually accurate answers. Quality evaluation techniques ensure the robustness and effectiveness of the synthetic RAFT dataset. The expert model for the multi-patent analysis task is fine-tuned on the synthetic dataset, enabling it to produce more accurate responses. Additionally, we construct a knowledge graph from patent documents to facilitate efficient semantic search and retrieval of relevant information. During inference, the expert model accesses the knowledge graph to provide auxiliary context, generating coherent, grounded, and accurate responses to complex patent-related queries.  In the following sections, we discuss the synthetic data generation process, including the creation of question-context-answer (QCA) triples, the Mixture-of-Agents (MoA) approach, and quality evaluation techniques. We also cover the integration of the knowledge graph and its impact on improving the expert model’s performance in patent analysis tasks during inference.

\vspace{-1mm}
\subsubsection{Synthetic Data Generation}
\label{subsec:Syn_datagen}
\vspace{-1mm}
Retrieval-Augmented Generation (RAG \cite{lewis2020retrieval}) provides LLMs with relevant external information from document databases, enabling them to generate outputs that are more contextually accurate, detailed, and grounded for ODQA tasks. This approach helps overcome the limitations of static, pre-trained knowledge in LLMs. In traditional RAG, documents are parsed and processed to extract text, then divided into smaller chunks using fixed-size chunking strategies to facilitate more precise retrieval of relevant content. Each chunk is embedded into a low-dimensional dense vector space that captures its semantic content, allowing for efficient indexing and retrieval of relevant chunks in response to queries. This method enhances generation by conditioning the language model on retrieved, contextually relevant chunks, leading to more accurate and grounded outputs. However, traditional RAG methods face several challenges that limit their effectiveness. These methods primarily rely on small text chunks, requiring the retriever to search through a large database to find relevant information. This can be inefficient, as the retriever often needs to recall numerous text chunks, sometimes necessitating re-ranking to optimize performance. Moreover, small text chunks can lead to semantic incompleteness and the loss of critical details due to document truncation. Dividing crucial context or concepts into multiple segments can impair coherence. Choosing an optimal chunk size is challenging: if chunks are too small, context is lost; if they are too large, retrieval becomes less precise—clearly, one size does not fit all. General-purpose LLMs are typically pre-trained on large text corpora using self-supervised learning techniques, such as predicting the next word in a sentence (autoregressive models) or filling in masked tokens (masked language models). To adapt LLMs for specific tasks, they undergo fine-tuning on task-specific datasets, enhancing their ability to follow instructions, improve contextual understanding, and solve complex problems. Despite these advancements, LLMs are generally not pre-trained or fine-tuned to inherently incorporate external retrieved context from databases, which is crucial for generating more accurate answers in ODQA. To address these limitations, the Retrieval-Augmented Fine-Tuning (RAFT \cite{zhang2024raft}) methodology optimizes LLMs to integrate retrieved content from external databases during fine-tuning. This approach enables language models to combine their internal parametric knowledge with dynamically retrieved external information, allowing for accurate and grounded responses in ODQA tasks. RAFT effectively overcomes the limitations of LLMs having only limited pre-trained, task-specific knowledge and their inability to utilize relevant external information from databases. However, synthetic RAFT datasets tailored for patent-related tasks are not readily available and require custom creation. To overcome these challenges, we employ a meticulous multi-step approach involving synthetic data generation and validation through a custom pipeline that creates high-quality RAFT datasets for fine-tuning LLMs to utilize relevant external information in patent-related ODQA tasks. It also filters out lower-quality instances to ensure robustness, employing both a `Reward-LLM-as-Judge` \cite{ouyang2022training} and `LLM-as-a-Judge` \cite{pan2024human} approach based on quality evaluation metrics. The RAFT dataset includes questions and relevant context retrieved from documents, along with chain-of-thought (CoT) reasoning-based answers.
For each text chunk of a patent document, we generate multiple questions using Gold-LLMs. We apply the Mixture-of-Agents (MoA \cite{wang2024mixture}) methodology to synthesize information from both the text chunks and relevant text chunks retrieved from patent databases by a retriever, generating contextually accurate answers to the questions. 
The MoA framework leverages the strengths of multiple Gold-LLMs within a layered architecture, enhancing natural language understanding and generation by incorporating outputs from other language models as additional context through collaborative synthesis. The MoA architecture consists of \(l\) layers, each containing \(n\) LLMs, denoted as \(A_{i,j}\) (i-th layer, j-th agent), which process inputs and generate outputs. These outputs are then aggregated and synthesized into a refined response \(y_i\), which serves as auxiliary input for the next layer. The output of the i-th layer, \(y_i\), is expressed as:

\vspace{-3mm}
\resizebox{0.925\linewidth}{!}{
\hspace{0mm}\begin{minipage}{\linewidth}
\begin{equation}
y_i = \bigoplus_{j=1}^n [A_{i,j}(x_i)] + x_1, \quad x_{i+1} = y_i \nonumber
\end{equation}
\end{minipage}
\vspace{-3mm}
}

where \(\bigoplus\) represents applying an aggregate-and-synthesize prompt to integrate multiple language model outputs into a coherent, high-quality response. In the final layer, a single, more capable Gold-LLM aggregates and synthesizes the outputs from the previous layers to generate a coherent, high-quality response. In the MoA technique, Gold-LLMs function either as proposers, generating diverse, context-rich responses, or as aggregators, synthesizing these into comprehensive outputs. Our implementation uses the LLaMA-3.1-405B-instruct and NemoTron-4-340B-instruct models as proposers, with GPT-4 Turbo serving as the aggregator. The architecture consists of two layers, each with two proposers and one aggregator working together to produce a coherent response. We utilize the `Reward-LLM-as-a-Judge', such as the multidimensional Nvidia NemoTron-4-340B-Reward model, to evaluate the quality of the question-answer pairs, focusing on attributes such as helpfulness, correctness, coherence, complexity, and verbosity. Additionally, we utilize `LLM-as-a-Judge', with GPT-4o acting as an expert judge to evaluate answers by rating them on relevance, accuracy, faithfulness, and coherence using a Likert scale from 1 to 5. At each layer, structured feedback from both the `Reward-LLM-as-a-Judge' and `LLM-as-a-Judge' models drives iterative refinement without fine-tuning, relying solely on prompting and generation to enhance response quality. In summary, the synthetic data generation pipeline operates by ingesting documents, generating diverse questions covering various scenarios and contexts, and using the MoA architecture to produce detailed, contextually relevant answers to fine-tune the expert model for multi-patent analysis. Our method then employs knowledge distillation to transfer knowledge from teacher models (LLaMA-3.1-405B-instruct, NemoTron-4-340B-instruct, GPT-4 Turbo) to a student model (GPT-4o mini), enabling the student model to replicate the larger model's outputs and behaviors through instruction-tuning on synthetic RAFT datasets. We fine-tune the expert model (i.e., GPT-4o mini) on the multi-patent analysis task using synthetic RAFT datasets generated from the MoA framework. During inference, the relevant context related to the end-user question is retrieved from the knowledge graph database (discussed in Subsection~\ref{subsec:Kg_semsearch_retrieval}), enhancing the expert model for multi-patent analysis response generation and resulting in coherent, knowledge-grounded, and accurate answers. In Subsection~\ref{subsec:Kg_semsearch_retrieval}, we discuss the construction of knowledge graphs from patents, followed by retrieval of relevant knowledge from the knowledge graph database, providing external context to the expert model for answering questions.

\vspace{-2mm}
\subsubsection{Knowledge Graph Modeling for Semantic Search and Retrieval}
\label{subsec:Kg_semsearch_retrieval}
\vspace{-1mm}
In Graph Retrieval-Augmented Generation (GRAG \cite{edge2024local, hu2024grag}), structured information is extracted from unstructured documents using parsing techniques and integrated into a knowledge graph. This facilitates efficient indexing and retrieval of contextually relevant content from knowledge graphs, enhancing language models for better performance in ODQA tasks. The process begins with parsing documents to extract structured data, such as text. This information is then integrated into graph databases such as Neo4j Aura, Amazon Neptune, or NebulaGraph, which are designed for storing and querying graph data. These databases organize the information into nodes and relationships within property graphs, preserving contextual and semantic information. The graph database enables efficient querying of property graphs, leading to more accurate responses compared to traditional keyword-based searches. Traditional RAG approaches rely on unstructured, plain-text retrieval, which has limitations when dealing with complex queries that require the integration of multiple, possibly disparate, text segments. In contrast, Graph RAG builds and utilizes a knowledge graph where entities and their relationships are explicitly modeled. This graph-based representation allows for better capturing and leveraging the underlying structure of the extracted data from documents, enhancing knowledge retrieval by enabling graph traversal structured by a schema and guided by an ontology. The vector semantic similarity search technique converts graph-structured data and queries into vector embeddings. Based on the similarity of these embeddings in the shared semantic space of the graph database, it retrieves and integrates nodes and relationships that are most relevant, enhancing retrieval accuracy by combining semantic similarity with contextual information. The retrieved information is then processed by a language model to generate accurate and contextually rich responses to complex queries. We lay the groundwork for constructing a knowledge graph by chunking text, embedding chunks, and extracting triples—data structures consisting of a subject, predicate, and object that represent relationships between entities—for efficient querying and retrieval. Text chunking involves breaking large documents into smaller, manageable segments, referred to as chunks, to enhance processing efficiency and retrieval accuracy. Using the sliding window technique, a fixed-size window moves across the text with a set stride, creating overlapping chunks that preserve context. Given a document \( D_{\text{doc}} \) with a total of \( N \) tokens, it is divided into overlapping chunks based on a chunk length \( l \) (the number of tokens in each chunk) and a stride \( s \). The \( i \)-th chunk \( c_i \) is defined as:

\vspace{-3mm}
\resizebox{0.925\linewidth}{!}{
\hspace{0mm}\begin{minipage}{\linewidth}
\begin{equation}
c_i = D_{\text{doc}}[(i-1) \times s : (i-1) \times s + l] \nonumber
\end{equation}
\end{minipage}
}

\( D_{\text{doc}}[(i-1) \times s : (i-1) \times s + l] \) represents the substring of \( D_{\text{doc}} \) starting at position \((i-1) \times s\) and extending for \( l \) tokens. 
Each chunk is associated with metadata, such as the title, summary, and keywords, which are concatenated with the text chunk itself to refine searches and provide contextual information. Text embedding models, such as OpenAI's text-embedding-3-small, convert these chunks into dense vectors \( \mathbf{e}_{c_i} \in \mathbb{R}^d \). This approach improves the preservation of contextual information and enhances the efficiency of vector search engines. The parsed text segments are stored as chunk nodes in a graph database. In the graph representation of text chunks, we denote nodes as \( V_T = \{v_{t_1}, v_{t_2}, \ldots, v_{t_K}\} \), where each node \( v_{t_i} \) corresponds to a specific chunk with associated content \( c_i \) and vector embeddings \( \mathbf{e}_{c_i} \). \( K \) represents the number of chunks generated from the document using the sliding window technique. \(|V_T|\) indicates that there are \( K \) nodes in the graph representation of text chunks. To facilitate easier querying, reasoning, and integration, these coarse-grained chunks are transformed into a fine-grained, structured, and standardized form. A Gold-LLM, such as GPT-4o, processes these chunk nodes to infer and extract knowledge graph triples. Knowledge graph triples \( (e_{s_{ij}}, r_{ij}, e_{o_{ij}}) \) are extracted from chunk nodes, where \( e_{s_{ij}} \) and \( e_{o_{ij}} \) are the subject and object nodes, respectively, and \( r_{ij} \) is the relation edge. Each text chunk \( c_i \) generates a set of triples \( \{ \tau_{i1}, \tau_{i2}, \ldots, \tau_{iM_i} \} \), where \( M_i \) is the number of triples extracted from chunk \( c_i \). Here, \( i \) is the index that identifies the specific chunk or node, and \( j \) is the index that distinguishes between multiple elements, such as triples, related to the same chunk or node. Each triple \( \tau_{ij} \) is defined as:

\vspace{-4mm}
\resizebox{0.925\linewidth}{!}{
\hspace{0mm}\begin{minipage}{\linewidth}
\begin{equation}
\tau_{ij} = (e_{s_{ij}}, r_{ij}, e_{o_{ij}}) \nonumber
\end{equation}
\end{minipage}
}

Here, \( \tau_{ij} \) represents the \( j \)-th extracted triple from chunk \( c_i \), with \( e_{s_{ij}} \) as the subject node, \( r_{ij} \) as the relation, and \( e_{o_{ij}} \) as the object node. The complete set of entity nodes, including both subject and object entities from all triples, is represented as:

\vspace{-1mm}
\resizebox{0.925\linewidth}{!}{
\hspace{0mm}\begin{minipage}{\linewidth}
\begin{equation}
V_E = \{e_{s_{ij}}, e_{o_{ij}} \mid \tau_{ij} = (e_{s_{ij}}, r_{ij}, e_{o_{ij}})\} \nonumber
\end{equation}
\end{minipage} 
}

These triples, consisting of subject, relation, and object entities formatted as single-hop paths, enhance the semantic understanding and usability of the information. Extracting triples provides finer granularity by breaking a text chunk into multiple distinct facts or relationships. Entity nodes are linked to chunk nodes through `MENTIONS` relationships, which signify that an entity is mentioned within a particular chunk. Let’s denote the `MENTIONS` relationship as \( \text{Ment}(v_{t_i}, e_{s_{ij}}) \) for a subject entity or \( \text{Ment}(v_{t_i}, e_{o_{ij}}) \) for an object entity. The dynamic ontology derived from the triples provides a framework for structured knowledge representation. This ontology describes node types (entity nodes \( V_E \)) and their interconnections (\( r_{ij} \), \( \text{Ment}(v_{t_i}, e_{s_{ij}}) \), and \( \text{Ment}(v_{t_i}, e_{o_{ij}}) \)), which together help in structuring and interpreting the knowledge contained within the graph. The schema outlines the database structure for organizing and representing entities (\( V_E \)) and relationships (\(r_{ij}, \text{Ment}(v_{t_i}, e_{s_{ij}}), \text{Ment}(v_{t_i}, e_{o_{ij}}) \)). In summary, GRAG extracts structured data from unstructured documents and integrates it into a graph database like Neo4j to enhance retrieval accuracy while maintaining context. Text is parsed and converted into nodes with metadata and vector embeddings, enabling structured knowledge representation and facilitating efficient semantic similarity searches. The KG expert model, such as GPT-4o mini, combines advanced language understanding capabilities with structured data stored in a graph database like Neo4j to interpret user queries for semantic similarity searches. We convert the user query \( Q \) into a vector embedding \( \mathbf{e}_Q \in \mathbb{R}^d \) using a sentence embedding model. To find relevant nodes, we compute the cosine similarity between \( \mathbf{e}_Q \) and all node vectors using \(\text{sim}(\mathbf{e}_Q, \mathbf{e}_{v_{x}}) = \frac{\mathbf{e}_Q \cdot \mathbf{e}_{v_{x}}}{\|\mathbf{e}_Q\| \|\mathbf{e}_{v_{x}}\|}\), where \( v_{x} \) is an entity node (\( e_{ij} \)). We then retrieve the top-\( K \) nodes with the highest similarity scores among entity nodes (\( e_{s_{ij}} \) and \( e_{o_{ij}} \)). For each selected entity node \( e_{ij} \), we perform graph traversal to extract one-hop triples: \(\mathcal{T}_{e_{ij}} = \{ (e_{ij}, r_{ij}, e_{o_{ij}}) \mid e_{o_{ij}} \text{ is \hspace{1mm} an \hspace{1mm} object \hspace{1mm} node \hspace{1mm} connected \hspace{1mm} to } e_{ij} \text{ \hspace{1mm} via \hspace{1mm} relation \hspace{1mm} } r_{ij} \}\) and retrieve the associated parent text chunk nodes (\( v_{t_i} \)). The parent chunk nodes containing both entities \( e_{ij} \) and \( e_{o_{ij}} \) are identified by \(\mathcal{C}_{e_{ij}} \hspace{1mm} \cap \hspace{1mm} \mathcal{C}_{e_{o_{ij}}} = \{ v_{t_i} \mid \text{chunk } v_{t_i} \text{ contains \hspace{1mm} both } e_{ij} \text{ and } e_{o_{ij}} \}\), while those containing either entity are found using \(\mathcal{C}_{e_{ij}} \hspace{1mm} \cup \hspace{1mm} \mathcal{C}_{e_{o_{ij}}} = \{ v_{t_i} \mid \text{chunk } v_{t_i} \text{ contains \hspace{1mm} either } e_{ij} \text{ or } e_{o_{ij}} \}\). By combining these results, we obtain:

\vspace{-2mm}
\resizebox{0.925\linewidth}{!}{
\hspace{0mm}\begin{minipage}{\linewidth}
\begin{equation}
\mathcal{R}_{e_{ij}} = \{ (e_{ij}, r_{ij}, e_{o_{ij}}, \mathcal{C}_{e_{ij}}, \mathcal{C}_{e_{o_{ij}}}) \mid (e_{ij}, r_{ij}, e_{o_{ij}}) \in \mathcal{T}_{e_{ij}} \} \nonumber
\end{equation}
\end{minipage}
}

When a query is posed, the knowledge graph is traversed to retrieve relevant nodes and relationships from the graph database and is integrated with the expert model's pre-existing knowledge to generate coherent and contextually appropriate responses. This approach leverages the expert model’s language generation capabilities while grounding its outputs in structured knowledge, leading to more accurate and informative answers. Entity deduplication in knowledge graphs resolves duplicates, reducing inconsistencies and redundancy. Our approach to entity deduplication integrates advanced techniques such as cosine similarity for vector-based evaluation and Levenshtein distance for string-based comparison, ensuring precise entity matching. For example, in a knowledge graph containing entities like `IBM' and `International Business Machines' or `Google LLC' and `Google', it first identifies these semantically related nodes and groups them into clusters based on overlapping characteristics. It then filters and merges these clusters, prioritizing the preservation of essential properties while eliminating redundancy. This process results in an optimized, deduplicated graph structure, enhancing the graph's effectiveness for knowledge extraction from unstructured documents and semantic search. The improved structure facilitates better contextual retrieval, leading to more accurate responses. In the specific domain of cross-document knowledge integration, a unified knowledge graph created from multiple patents, supported by LLM-driven ontology extraction and schema alignment, effectively organizes and standardizes patent data, enabling precise exploration, easier identification of trends and gaps, and promoting faster, data-driven innovation. In summary, transforming documents into structured knowledge graphs enables efficient, context-aware retrieval of information from unstructured data, enhancing the framework's ability to reason and generate relevant solutions for complex patent-related tasks, thereby improving efficiency in Q\&A tasks. Please note: the expert model for multi-patent analysis accesses the KG search engine as a tool.

\vspace{-1mm}
\subsection{Scientific Hypothesis Generation}
\vspace{-1mm}
In this work, we employed teacher-student transfer learning via knowledge distillation using Gold-LLMs, such as GPT-4o, to extract or generate scientific hypotheses from granted patents. This process created a synthetic dataset to fine-tune expert models, such as GPT-4o-mini, for the task of patent hypothesis generation. It is important to note that hypothesis generation provides a broader, conceptual explanation of the problem the invention solves and how it works, rather than focusing on legal protection. Patent claims, by contrast, define the specific legal boundaries and technical implementations that the patent protects, emphasizing the structural or procedural aspects of the invention in a precise and formal manner. We utilized Gold-LLMs, such as GPT-4o, to analyze a set of granted patents, extracting Subject-Action-Object (SAO) triplets to uncover the key hypotheses and innovations within the patent documents. The approach began by parsing the granted patents, focusing on key sections such as claims and detailed descriptions, which outline the scope of the patent’s innovation. The Gold-LLM identified and extracted SAO triplets, which are essential for identifying the central technical contributions of the patents. In this context, the subject represents the core invention or technology, the action describes the method or process being applied, and the object defines the outcome or product of the action. The Gold-LLM then synthesized the SAO triplets for each patent, generating a comprehensive hypothesis that represents the innovations and technical contributions of the patents. This hypothesis provides a high-level overview of the key technologies and innovations within each patent, allowing for the identification of unique contributions, and potential advancements. By extracting a hypothesis for each patent, it becomes possible to cross-reference these insights with prior art and technical literature, ensuring that the innovations are novel and consistent with the granted claims. This process enables scalable, automated analysis of patents, streamlining the detection of valuable technological breakthroughs and trends in patent data. We used standard natural language processing (NLP) metrics such as BLEU, ROUGE, and METEOR to compare the generated hypotheses from expert models against reference summaries from Gold LLMs. Table \ref{tab:universal_transformer_sao} illustrates the hypothesis generation for the Universal Transformers \cite{US10740433B2} patent using a Gold LLM such as GPT-4o.

\vspace{-2mm}
\begin{table*}[ht!]
\begin{tcolorbox}[colback=white, colframe=black, coltitle=black, colbacktitle=white, title= \textbf{Subject-Action-Object (SAO) Structure and Hypothesis for Universal Transformer}.]
\footnotesize
\vspace{-1mm}
\begin{tcolorbox}[colback=white, colframe=black, coltitle=black, colbacktitle=white, title=\textbf{Subject.}] 
\vspace{-1mm}
\textbf{Universal Transformer (UT) architecture.}
\vspace{-2mm}
\end{tcolorbox}
\vspace{-1mm}
\begin{tcolorbox}[colback=white, colframe=black, coltitle=black, colbacktitle=white, title=\textbf{Action.}]
\vspace{-1mm}
\textbf{Introduces recurrent self-attention across depth, parameter sharing across layers, and an adaptive computation time (ACT) mechanism to dynamically adjust the number of computational steps per sequence element.}
\vspace{-2mm}
\end{tcolorbox}
\vspace{-1mm}
\begin{tcolorbox}[colback=white, colframe=black, coltitle=black, colbacktitle=white, title=\textbf{Object.}]
\vspace{-1mm}
\textbf{Overcome the limitations of traditional Transformers and LSTMs, particularly in generalization to longer sequences and computational efficiency in tasks like machine translation, language modeling, and algorithmic problem-solving, resulting in state-of-the-art performance on a variety of complex sequence tasks.}
\vspace{-2mm}
\end{tcolorbox}
\vspace{-1mm}
\begin{tcolorbox}[colback=white, colframe=black, coltitle=black, colbacktitle=white, title=\textbf{Hypothesis.}]
\vspace{-1mm}
\textbf{The Universal Transformer (UT) architecture introduces a novel combination of recurrent self-attention mechanisms in depth with parameter sharing across layers and an adaptive computation time (ACT) mechanism for per-position dynamic adjustment of computational depth. Unlike traditional Transformers, which fix the depth of computation per sequence element, or LSTMs, which rely on sequential recurrence, the UT’s recurrence in depth allows iterative refinement of representations and dynamic computation, enabling improved generalization to longer sequences and complex algorithmic tasks. This innovation offers computational universality and superior performance on sequence-to-sequence tasks like machine translation and language modeling, where it consistently surpasses both feed-forward Transformers and LSTMs.}
\vspace{-2mm}
\end{tcolorbox}
\vspace{-3mm}
\end{tcolorbox}
\vspace{-3mm}
\caption{The table provides the Subject-Action-Object (SAO) structure and hypothesis for the Universal Transformer (UT) architecture, highlighting its key innovations and advantages over traditional models like Transformers and LSTMs.}
\label{tab:universal_transformer_sao}
\vspace{-2mm}
\end{table*}


\end{document}